\documentclass[10pt,twocolumn,letterpaper]{article}

\usepackage{float}
\usepackage[pagenumbers]{cvpr}

\newcommand{\TODO}[1]{\textbf{\color{red}[TODO: #1]}}
\usepackage{multirow}
\usepackage{colortbl}
\usepackage{float}
\usepackage[table]{xcolor}
\usepackage[nosfdefault]{comicneue} 
\usepackage[T1]{fontenc}
\usepackage{booktabs}
\usepackage{placeins}
\usepackage{multirow}
\usepackage{makecell}
\usepackage[most]{tcolorbox}
\usepackage{fancybox}
\usepackage{tikz}
\usepackage{caption}

\definecolor{oursbg}{RGB}{245, 249, 255}
\definecolor{oursheaderbg}{RGB}{228, 238, 255}
\definecolor{oursaccent}{RGB}{40, 90, 170}

\newcommand{\oursrow}{\rowcolor{oursbg}}
\newcommand{\oursblock}{\rowcolor{oursheaderbg}}
\newcommand{\best}[1]{\textbf{\textcolor{oursaccent}{#1}}}
\newcommand{\ourslayers}[1]{\textbf{#1 interaction layers}}

\newcommand{\tabcomic}{\fontfamily{ComicNeue-TLF}\selectfont}

\renewcommand{\TODO}[1]{}

\definecolor{cvprblue}{rgb}{0.21,0.49,0.74}
\definecolor{darkgreen}{RGB}{25,125,32}

\usepackage[pagebackref,breaklinks,colorlinks,allcolors=cvprblue]{hyperref}

\def\paperID{4602} 
\def\confName{CVPR}
\def\confYear{2026}

\title{Bridging Hidden States in Vision–Language Models}

\author{Benjamin Fein-Ashley\\
University of Southern California\\
\and
Jacob Fein-Ashley\\
University of Southern California\\
}

\begin{document}
\maketitle
\begin{abstract}
Vision-Language Models (VLMs) are a new family of models that align image content with natural language. Existing approaches typically fuse either (a) early: by mixing tokens/features inside the encoders, or (b) late: by comparing pooled embeddings. Many methods also tie fusion to an autoregressive decoder. However, the hidden states of both modalities already carry rich, modality-specific structure (spatial layout in vision; syntax and semantics in text), so directly aligning these states is a natural way to match what the two modalities ``think''. We propose a lightweight fusion module: a few cross-only, bidirectional attention layers placed near the top of both encoders. Each layer projects the vision and text encoder hidden-state sequences into a shared space, attends across modalities, and sends gated residual updates back, with simple stabilizers to improve alignment. The encoders remain non-causal and strong for understanding, while generation stays cleanly decoupled via an optional decoder. Across standard retrieval, VQA, and visual reasoning benchmarks, BRIDGE outperforms comparable VLMs while preserving the bi-encoder efficiency of contrastive models. We make our code publicly available at \url{https://github.com/jfeinashley/BRIDGE}.
\end{abstract}    
\section{Introduction}
\label{sec:intro}
Vision–Language Models (VLMs) bring visual understanding to Large Language Models (LLMs). Building on advances in language modeling such as GPT-4~\cite{gpt4} and recent multimodal systems (e.g.,~Gemini~\cite{Gemini} and Sonnet 4.5~\cite{sonnet45}), interest in scalable multimodal reasoning continues to grow. Most VLMs follow a modular design with three components: a \textit{vision encoder}, a \textit{text encoder}, and a \textit{fusion module} (e.g.,~\cite{LXMERT, VisualBERT, ViLT, CLIP, ALIGN}). The vision encoder is commonly a transformer (ViT)~\cite{ViT} that captures spatial structure via patches or regions; the text side is a pretrained transformer/LLM. Fusion integrates the two modalities, and an optional decoder (often an LLM) handles generation for captioning or other objectives. VLMs typically adopt one of two broad fusion strategies:
\begin{itemize} 
\item \textbf{Early fusion}: mixing vision and text tokens/features inside a shared module (e.g., concatenating image patch tokens and word tokens and feeding them into a joint transformer).

\item \textbf{Late fusion}: encoding vision and text separately, then aligning their pooled embeddings via contrastive or matching losses.
\end{itemize}
Each approach has trade-offs: early fusion enables tight interactions but may blur unimodal specialization; late fusion preserves strong unimodal backbones but can limit fine-grained cross-modal reasoning~\cite{Yuksekgonul}. To capture richer interactions beyond contrastive-only alignment, prior work combines contrastive, matching, and captioning objectives~\cite{ALBEF, CoCa}.

\begin{figure*}
    \includegraphics[width=\textwidth]{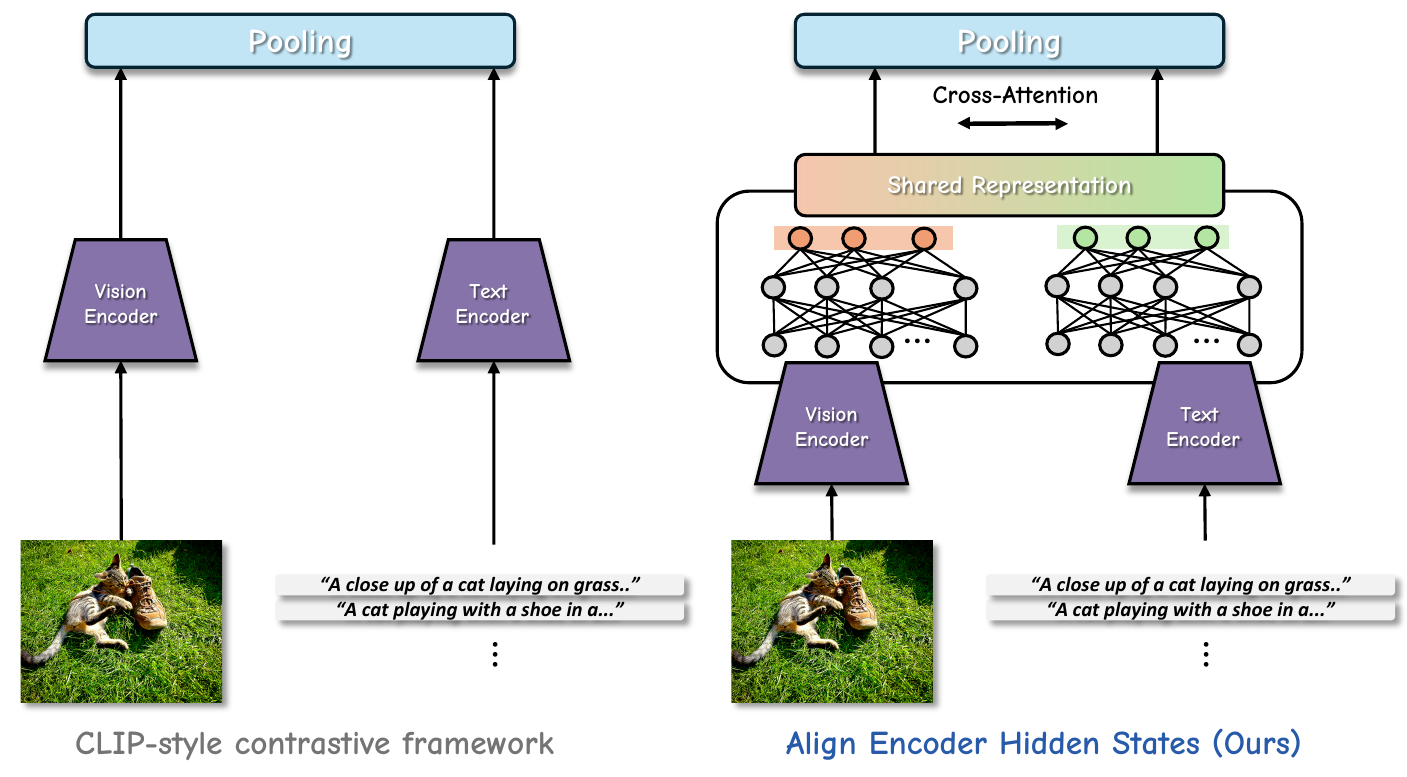}
    \caption{Visual demonstration of BRIDGE alignment space}
\end{figure*}

Our method, \textbf{BRIDGE}, is a lightweight middle ground that adds a few cross-only, bidirectional attention layers near the top of the vision and text encoders to exchange information in a shared latent space while preserving efficient bi-encoder inference.

This work makes the following contributions:
\begin{itemize}
    \item We introduce BRIDGE, a lightweight fusion module that inserts cross-only, bidirectional attention layers near the top of vision and text encoders to align full hidden-state sequences in a shared latent space with gated residual updates.
    \item We propose a simple training recipe that combines standard contrastive and matching objectives with a cycle-consistent cross-attention loss to explicitly encourage stable bidirectional alignment between image and text tokens.
    \item We show that BRIDGE preserves bi-encoder efficiency for retrieval while improving performance on standard retrieval, VQA, and visual reasoning benchmarks compared to comparable baselines.
\end{itemize}
\section{Related Works}
\label{sec:related}

\subsection{Vision-Language Pretraining}
Recent advances in VLMs have been propelled by multi-modal pretraining. Early approaches in vision-language pretraining models (e.g. LXMERT \cite{LXMERT}, VisualBERT \cite{VisualBERT}, UNITER \cite{UNITER}) use a transformer-based architecture with cross-attention, relying on pretrained object detectors to fuse image features with text tokens. Foundational contrastive approaches such as CLIP ~\cite{CLIP} and ALIGN \cite{ALIGN} independently encode images and text, aligning pooled embeddings in a shared space. These late-fusion approaches allow for strong zero-shot generalization by aligning text and image embeddings through contrastive learning, but have limited cross-modal reasoning out of distribution \cite{Yuksekgonul}. \\
\paragraph{Multi-Objective Learning.}
Many models take an intermediate approach. CoCa \cite{CoCa} unifies contrastive objectives along with captioning. Another approach is cross-attention between modalities. BLIP \cite{BLIP} uses fine tuning with cross-attention to align modalities, while BLIP-2 \cite{BLIP2} introduces the querying transformer (Q-former) to bridge the two encoders. ALBEF \cite{ALBEF} integrates contrastive and matching objectives along with cross-attention. This combined approach is key for capturing multi-modal interactions. \\
\paragraph{MLLMs and Instruction Tuning.}
VLMs are often extended to visual question-answering (VQA) and reasoning tasks by adding an LLM as a decoder. Early systems such as Flamingo~\cite{Flamingo} and PaLI~\cite{pali} incorporate cross-attention to integrate image features with language, enabling open-ended visual reasoning. 
Instruction tuning has emerged as a key strategy for aligning multimodal capabilities. Models such as InstructBLIP~\cite{instructblip}, LLaVA~\cite{llava}, and MiniGPT-4~\cite{minigpt4} fine-tune pretrained VLM backbones using image–text–instruction triplets. More recent instruction-tuned MLLMs, such as Qwen-VL~\cite{qwenvl}, InternVL~\cite{internvl}, and Kosmos-2~\cite{kosmos2}, scale with larger training corpora and more powerful LLM backbones, achieving high performance on VQA tasks. Many models rely on LLM self-attention for alignment, with image and text tokens concatenated and used as the input sequence for the LLM~\cite{blip3,llava}. This constrains token-level grounding and limits detailed bidirectional reasoning between modalities. These limitations motivate architectures that unify scalable LLM reasoning with deeper and more efficient multimodal fusion.

\subsection{Cross-Modal Representation}
Novel alignment methods and representation learning are of interest to overcome these limitations. Query-based alignment, introduced in BLIP-2's Q-former \cite{BLIP2}, has been widely adopted. Qwen2-VL uses M-RoPE encoding to improve the fusion of positional information between modalities \cite{Qwen2VL}. Adapter-based fusion has become a very active approach, with models inserting small trainable modules into frozen encoder backbones without the need for full fine-tuning \cite{VLAdapter, PalmV2, CrossModalAdapter}. Many of these approaches focus on parameter efficient tuning at a small or negligible performance cost. 
Our approach provides a scalable fusion strategy with a new approach to modality cross-attention. Hidden states of both the text and vision encoder \textit{directly} interact, removing the need for additional adapter stacks. Our architecture allows for scalable overhead and reduces the reliance on LLM self-attention for cross-modal representation in VQA and reasoning tasks. 
\section{Method}
\label{sec:method}
We propose \textbf{BRIDGE} (\underline{b}idi\underline{r}ectional h\underline{i}\underline{d}den-state \underline{e}xchan\underline{g}e), a framework to directly align hidden states in text and vision encoders for greater multi-modal understanding. This section introduces our model architecture, enumerates learning objectives, and outlines the modular approach for task-specific tuning.
\subsection{Model Architecture}
Our architecture begins with multi-modal streams entering respective frozen encoders. We apply cross-attention on a select number of interaction layers at the top of each encoder to align their hidden states in a shared space. Interaction layers in the encoders are unfrozen in staged training.
\paragraph{Encoders.} We employ vision and text encoders. Given image patches and tokenized text, the encoders produce hidden states $H_t^{(l)}$ and $H_v^{(l)}$ at each layer $l$. \\
\textbf{Hidden State Extraction.} $H_t^{(l)}$ and $H_v^{(l)}$ from the encoders are regularized and aggregated with reverse-projected outputs from gating. We use layer normalization and project the hidden states to the interaction layers.\\
\textbf{Interaction layers.} Cross-only multi-head attention (MHA) is applied on the projected hidden states. There is no self-attention; vision only attends to text and text only attends to vision. Text and vision attention are projected to a shared dimension $d_s$. After dropout and output projection, outputs are gated with learnable parameters. This provides enhanced modality fusion with adaptive contribution of text and vision, improving stability and performance. \\
\textbf{Downstream Tasks.} InfoNCE loss~\cite{CLIP} promotes unimodal representation, allowing for fast retrieval without calculating cross-attention during inference. Optionally, a decoder (an LLM in our case) can be attached after pooling for VQA and reasoning tasks.

\paragraph{Interaction block (one layer).}
Given vision and text hidden states
$H_v^{(l)} \in \mathbb{R}^{N_v \times d_v}$ and
$H_t^{(l)} \in \mathbb{R}^{N_t \times d_t}$,
we first map them to a shared space of width $d_s$ with PreNorm:
\begin{align}
Z_v^{(l)} &= \mathrm{LN}\!\big(H_v^{(l)}\big)\, W_{v\to s}, &
Z_t^{(l)} &= \mathrm{LN}\!\big(H_t^{(l)}\big)\, W_{t\to s}. \label{eq:proj-shared}
\end{align}
We then apply \emph{cross-only} multi-head attention in both directions
(standard dot-product MHA, but with queries from one modality and
keys/values from the other):
\begin{align}
A_t^{(l)} &= \mathrm{MHA}_{\text{cross}}\!\big(Z_t^{(l)}, Z_v^{(l)}\big), \nonumber\\
A_v^{(l)} &= \mathrm{MHA}_{\text{cross}}\!\big(Z_v^{(l)}, Z_t^{(l)}\big). \label{eq:mha-cross}
\end{align}
and update the encoders with a gated residual in their native spaces:
\begin{align}
H_t^{(l+1)} &= H_t^{(l)} + g_t^{(l)}\, A_t^{(l)} W_{s\to t}, \nonumber\\
H_v^{(l+1)} &= H_v^{(l)} + g_v^{(l)}\, A_v^{(l)} W_{s\to v}, \label{eq:gated-residual-main}
\end{align}
where $g_t^{(l)}, g_v^{(l)} \in (0,1)$ are learned scalar gates
and $W_{v\to s}, W_{t\to s}, W_{s\to t}, W_{s\to v}$ are learned projections.
We stack $Q$ such interaction blocks near the top of both encoders.

\begin{figure*}[tp]
    \includegraphics[width=\textwidth]{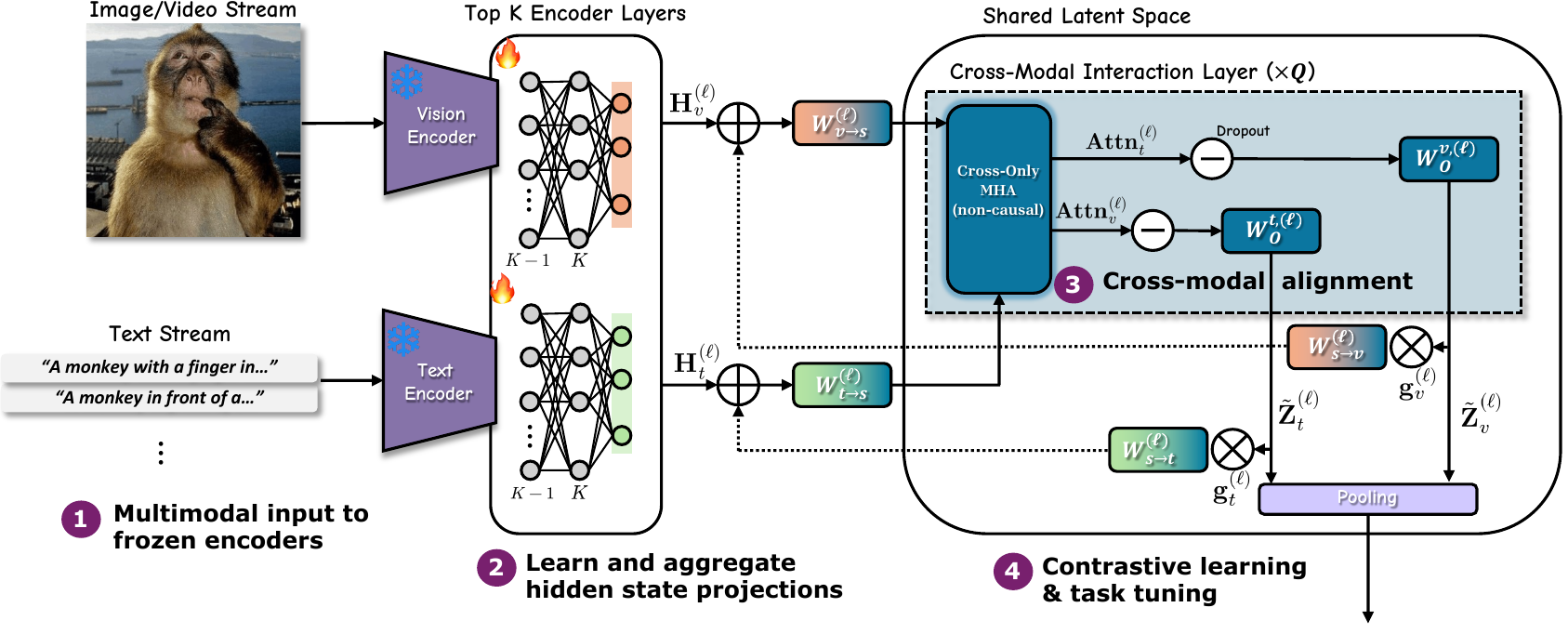}
    \caption{Architecture framework for \textbf{BRIDGE}. We propose an architecture where the hidden states of text and vision encoders are aligned directly rather than through pooled embeddings and contrastive loss. In a shared latent space, cross-only MHA is applied with residuals reverse-projected to respective embedding spaces.}
\end{figure*}

\subsection{Learning Objectives}
We jointly optimize standard VLM objectives (image-text contrastive (ITC) loss, masked image/language modeling) and two losses designed to promote cross-modal representation. We utilize a linear combination of losses with learnable scaling parameters \\
\textbf{(1) Image-Text Matching (ITM) with semi-hard negatives.}
ITM loss is used to activate the text encoder. We propose a binary classification task using semi-hard negatives. A small multi-layer perceptron (MLP) on cross-modal embeddings produces a logit indicating contrastive similarity.\\
\textbf{(2) Cycle-consistent cross-attention.} 
This objective promotes stable alignment by enforcing cycle-consistent cross-attention. 
\paragraph{Training objective.}
From the cross-fused encoders we obtain pooled text and image embeddings
$p_t^{(i)}, p_v^{(i)}$ for each paired example $i$ in a minibatch of size $B$.
Following CLIP~\cite{CLIP}, we use a symmetric InfoNCE image--text contrastive loss
with similarities $S_{ij} = \langle p_t^{(i)}, p_v^{(j)} \rangle / \tau_c$:
\begin{align}
\mathcal{L}_{\mathrm{itc}} &=
-\frac{1}{2B} \sum_{i=1}^{B}
\log \frac{\exp(S_{ii})}{\sum_{j=1}^{B} \exp(S_{ij})} \nonumber\\
&\quad
-\frac{1}{2B} \sum_{j=1}^{B}
\log \frac{\exp(S_{jj})}{\sum_{i=1}^{B} \exp(S_{ij})}. \label{eq:itc-main}
\end{align}
For cycle-consistent cross-attention, we average the per-head attention
probabilities from text-to-vision and vision-to-text within each interaction layer
to obtain $\bar{P}_{t\to v}$ and $\bar{P}_{v\to t}$. The round-trip products
\[
C_v = \bar{P}_{t\to v}\, \bar{P}_{v\to t}, \qquad
C_t = \bar{P}_{v\to t}\, \bar{P}_{t\to v}
\]
are encouraged to be close to identity via a simple diagonal penalty:
\begin{align}
\mathcal{L}_{\mathrm{cyc}} &=
-\frac{1}{2}\Big(
\operatorname{mean}\big(\log \operatorname{diag}(C_v)\big) \nonumber\\
&\qquad\quad
+\operatorname{mean}\big(\log \operatorname{diag}(C_t)\big)
\Big). \label{eq:cyc-main}
\end{align}
In addition, we use a binary image--text matching loss $\mathcal{L}_{\mathrm{itm}}$
with semi-hard negatives, and optional masked language/image modeling losses
$\mathcal{L}_{\mathrm{mlm}}$ and $\mathcal{L}_{\mathrm{mim}}$.
The total objective is a weighted sum of all of the listed loss functions.

\subsection{Task-Specific Tuning}
The proposed framework supports flexible downstream adaptation. For discriminative tasks such as VQA, we fine-tune the cross-fused encoder outputs $\bar{H}_t$ and $\bar{H}_v$ with an LLM decoder. For retrieval benchmarks, we use the unimodal projection heads to enable efficient similarity search without cross-attention at inference. The combination of unimodal and cross-modal ITC loss promotes unimodal representation for retrieval.

This modular design allows the model to serve as both a high-capacity vision–language encoder and a foundation for generative multi-modal reasoning tasks, providing a unified architecture for alignment, retrieval, and understanding tasks.

\subsection{Training Scheme}
We find that it is best to train our method in three steps. \textbf{Stage A (stabilize)}: freeze both encoders; train only interaction layers, gates, positional biases, and pooled heads. \\
\textbf{Stage B (align)}: unfreeze top $K$ blocks of each encoder (closest to interaction layers). \\
\textbf{Stage C (task tuning)}: Involves finetuning for specific tasks. For understanding tasks, attach a classifier head on pooled $\bar{H}_t$/$\bar{H}_v$ or token-wise heads on $\bar{H}_t,\bar{H}_v$. For retrieval, use $P_t^u,P_v^u$ for fast bi-encoder search (no extra fine-tuning needed). For generation, attach a separate causal decoder that cross-attends to frozen vision features or compact bridge latents.

\section{Experiments}
\label{sec:experiments}

All models and configurations that we display in these experiments portray the maximum limit of our computational budget: that is, we scale the models as much as we can to have a fair comparison. Additionally, we compare against works/models that have a similar parameter size and training/pretraining configuration to exhibit the most ``apples-to-apples'' comparison that is possible.

\subsection{Image--Text Retrieval}

We evaluate BRIDGE on bidirectional image--text retrieval using the MSCOCO dataset~\cite{lin2014coco} with the standard Karpathy 5K split~\cite{karpathy2015} and the Flickr30K dataset~\cite{flickr}. All models are pretrained on Visual Genome~\cite{krishna2017visualgenome}, Conceptual Captions~\cite{sharma2018conceptual}, and Conceptual 12M~\cite{changpinyo2021cc12m}. We then fine-tune on the union of the MSCOCO train and restval splits for COCO retrieval, and on the standard train split for Flickr30K.

\begin{table*}[h]
\centering
\small
\setlength{\tabcolsep}{6pt}
\renewcommand{\arraystretch}{1.1}

\begin{tikzpicture}
\node[inner sep=2.5pt] (tbl) {%
  {\tabcomic
  \begin{tabular}{lccrcccccccc}
    \multirow{2}{*}{Model} &
    \multicolumn{2}{c}{Backbone} &
    \multirow{2}{*}{\# Params} &
    \multicolumn{4}{c}{MSCOCO (Karpathy 5K)} &
    \multicolumn{4}{c}{Flickr30K (1K test)} \\
    \cmidrule(lr){2-3} \cmidrule(lr){5-8} \cmidrule(lr){9-12}
    & Image & Text & &
    TR@1 & TR@5 & IR@1 & IR@5 &
    TR@1 & TR@5 & IR@1 & IR@5 \\
    \midrule
    CLIP~\cite{CLIP}       & ViT-B/32 & Transformer & 151M &
      37.8 & 62.4 & 58.4 & 81.5 &
      86.5 & 98.0 & 67.0 & 88.9 \\
    ALBEF~\cite{ALBEF}     & ViT-B/16 & BERT-Base   & 203M &
      77.6 & 94.3 & 60.7 & 84.3 &
      77.6 & 94.1 & 61.0 & 84.5 \\
    BLIP (14M)~\cite{BLIP} & ViT-B/16 & BERT-Base   & 213M &
      80.6 & 95.2 & 63.1 & 85.3 &
      96.9 & 99.9 & 87.5 & 97.6 \\
    \addlinespace[3pt]
    \oursblock
    \multicolumn{12}{l}{\textbf{BRIDGE (Ours)}} \\
    \oursrow
    \ourslayers{2} & ViT-B/16 & BERT-Base & 236M &
      \textit{81.3} & \textit{96.3} & \textit{66.9} & \textit{86.4} &
      \textit{97.2} & \textit{99.9} & \textit{88.2} & \textit{97.8} \\
    \oursrow
    \ourslayers{4} & ViT-B/16 & BERT-Base & 250M &
      \textit{81.5} & \textit{96.5} & \textit{67.2} & \textit{86.7} &
      \textit{97.4} & \textit{99.9} & \textit{88.5} & \textit{97.9} \\
    \oursrow
    \ourslayers{6} & ViT-B/16 & BERT-Base & 264M &
      \best{\textit{81.6}} & \best{\textit{96.6}} &
      \best{\textit{67.5}} & \best{\textit{86.9}} &
      \best{\textit{97.5}} & \best{\textit{99.9}} &
      \best{\textit{88.8}} & \best{\textit{98.0}} \\
  \end{tabular}
  }%
};

\draw[line width=0.08pt, rounded corners=4pt]
  (tbl.south west) rectangle (tbl.north east);

\draw[line width=0.08pt, rounded corners=4pt, opacity=0.6]
  ([xshift=0.5pt,yshift=0.5pt]tbl.south west)
    rectangle
  ([xshift=0.5pt,yshift=0.5pt]tbl.north east);
\end{tikzpicture}

\caption{\textbf{Image--Text Retrieval on MSCOCO and Flickr30K.}
Comparison of recent VLMs on the MSCOCO Karpathy 5K split~\cite{karpathy2015} and the Flickr30K 1K test set~\cite{flickr}.
TR: text-to-image retrieval; IR: image-to-text retrieval. All values are Recall (\%).}
\label{tab:retrieval}
\end{table*}
For each image--caption pair, we optimize the unimodal and cross-modal InfoNCE objectives together with the image--text matching (ITM) loss and the cycle-consistent cross-attention loss introduced in Section~\ref{sec:method}. At inference time, BRIDGE performs retrieval using only the unimodal projection heads from the text and vision encoders, without running cross-attention, keeping the test-time cost comparable to contrastive models such as CLIP.

We report Recall@K for text-to-image retrieval (TR@1/5) and image-to-text retrieval (IR@1/5), following prior work. As shown in Table~\ref{tab:retrieval}, BRIDGE consistently improves over CLIP, ALBEF, and BLIP on MSCOCO when using comparable ViT-B/16 and BERT-Base backbones. Increasing the number of interaction layers strengthens cross-modal alignment and yields monotonic gains in both TR and IR performance, with the 6-layer variant achieving the best overall R@1 and R@5 under the same pretraining data. Table~\ref{tab:retrieval} also reports results on Flickr30K, demonstrating that the same unimodal retrieval heads generalize to a distinct retrieval benchmark.

\subsection{Visual Question Answering (VQA)}
\label{subsec:vqa}

We further evaluate BRIDGE on open-ended visual question answering using the VQAv2 benchmark~\cite{VQA}, which augments MSCOCO images with balanced question--answer pairs to reduce language priors. Following the standard protocol, we train on the official train and validation splits and report single-model performance on the test-dev and test-std splits via the evaluation server.

\textbf{Setup.}
For VQA, we attach an autoregressive Transformer decoder on top of the cross-fused representations $\bar{H}_t$ and $\bar{H}_v$ introduced in Section~\ref{sec:method}. The decoder is initialized from a pretrained LLM and fine-tuned jointly with the interaction layers, while keeping the unimodal encoders frozen except for the selected cross-attention blocks. Questions are tokenized and concatenated with a special \texttt{[VQA]} token, and answers are generated as short text sequences. We follow prior work~\cite{UNITER,OSCAR,VinVL,ALBEF,BLIP,SimVLM} and restrict the answer space to the most frequent training answers, optimizing a token-level cross-entropy loss over the decoded sequence, i.e., a standard language modeling (LM) loss. In contrast, UNITER~\cite{UNITER}, OSCAR~\cite{OSCAR}, and VinVL~\cite{VinVL} formulate VQA as answer classification and apply a linear classifier on top of the multimodal \texttt{[CLS]} representation over the same fixed answer set without an autoregressive decoder.

\textbf{Baselines.}
We compare BRIDGE against strong vision--language pretraining methods spanning both region-based and fully end-to-end architectures. Region-level models such as UNITER~\cite{UNITER}, OSCAR~\cite{OSCAR}, and VinVL~\cite{VinVL} rely on Faster R-CNN features and Transformer-based fusion, while more recent patch-based VLP models including ALBEF~\cite{ALBEF}, BLIP~\cite{BLIP}, and SimVLM~\cite{SimVLM} use ViT-style encoders and large-scale weakly supervised pretraining. All baselines are evaluated under their published single-model settings with comparable backbone capacity.

\textbf{Results.}
Table~\ref{tab:vqa} summarizes VQAv2 performance. Region-based methods achieve strong accuracy but require an external detector, while patch-based models (ALBEF, BLIP, SimVLM) benefit from tighter image--text coupling. BRIDGE builds on a ViT-B/16 image encoder and a BERT-Base text encoder, but differs in that it explicitly aligns hidden states via bidirectional cross-only attention. Across different numbers of interaction layers, BRIDGE provides competitive VQA performance while maintaining the same unimodal encoders used for retrieval in Section~\ref{sec:experiments}. Increasing the number of interaction layers generally improves accuracy, reflecting the benefit of deeper cross-modal fusion for compositional reasoning.

\begin{table}
\centering
\small
\setlength{\tabcolsep}{5pt} 
\renewcommand{\arraystretch}{1.2}

\resizebox{\columnwidth}{!}{%
\begin{tikzpicture}
\node[inner sep=2.5pt] (tbl) {%
  {\tabcomic
  \begin{tabular}{lccccc}
    \multirow{2}{*}{Model} &
    \multicolumn{2}{c}{Backbone} &
    \multicolumn{2}{c}{VQAv2~\cite{VQA}} \\
    \cmidrule(lr){2-3} \cmidrule(lr){4-5}
    & Image & Text & test-dev & test-std \\
    \midrule
    UNITER~\cite{UNITER}      & Faster R-CNN & BERT-Base   & 73.8 & 74.0 \\
    OSCAR~\cite{OSCAR}        & Faster R-CNN & BERT-Base   & 73.6 & 73.8 \\
    VinVL~\cite{VinVL}        & Faster R-CNN & BERT-Base   & 76.5 & 76.6 \\
    ALBEF~\cite{ALBEF}        & ViT-B/16     & BERT-Base   & 75.8 & 76.0 \\
    BLIP (14M)~\cite{BLIP}    & ViT-B/16     & BERT-Base   & 78.3 & 78.3 \\
    SimVLM~\cite{SimVLM}      & Transformer  & Transformer & 80.0 & 80.3 \\
    \addlinespace[3pt]
    \oursblock\oursrow
    BRIDGE (Ours) & ViT-B/16 & BERT-Base &
      \best{\textit{80.6}} & \best{\textit{80.7}} \\
  \end{tabular}
  }%
};

\draw[line width=0.08pt, rounded corners=4pt]
  (tbl.south west) rectangle (tbl.north east);

\draw[line width=0.08pt, rounded corners=4pt, opacity=0.6]
  ([xshift=0.5pt,yshift=0.5pt]tbl.south west)
    rectangle
  ([xshift=0.5pt,yshift=0.5pt]tbl.north east);
\end{tikzpicture}
}

\caption{\textbf{VQA on VQAv2.} Comparison of BRIDGE with prior vision--language models on the VQAv2 benchmark~\cite{VQA}. All values are overall VQA accuracy (\%).}
\label{tab:vqa}
\end{table}

\subsection{Natural Language Visual Reasoning}
\label{subsec:nlvr2}

We evaluate BRIDGE on natural language visual reasoning using the NLVR2 benchmark~\cite{NLVR2}, which requires deciding whether a natural language statement correctly describes a pair of images. We fine-tune a lightweight binary classifier on top of the cross-fused representation from Section~\ref{sec:method}, reusing the same ViT-B/16 image encoder and BERT-Base text encoder as in our retrieval and VQA experiments. 
\newpage
Table~\ref{tab:nlvr2} reports accuracy on the NLVR2 dev and public test (Test-P) splits.
\begin{table}[tp]
\centering
\small
\setlength{\tabcolsep}{5pt} 
\renewcommand{\arraystretch}{1.2}

\resizebox{\columnwidth}{!}{%
\begin{tikzpicture}
\node[inner sep=2.5pt] (tbl) {%
  {\tabcomic
  \begin{tabular}{lccccc}
    \multirow{2}{*}{Model} &
    \multicolumn{2}{c}{Backbone} &
    \multicolumn{2}{c}{NLVR2~\cite{NLVR2}} \\
    \cmidrule(lr){2-3} \cmidrule(lr){4-5}
    & Image & Text & dev & test-P \\
    \midrule
    ALBEF (4M)~\cite{ALBEF}    & ViT-B/16 & BERT-Base & 80.24 & 80.50 \\
    ALBEF (14M)~\cite{ALBEF}   & ViT-B/16 & BERT-Base & 82.55 & 83.14 \\
    TCL~\cite{TCL}             & ViT-B/16 & BERT-Base & 80.54 & 81.33 \\
    BLIP (14M)~\cite{BLIP}     & ViT-B/16 & BERT-Base & 82.67 & 82.50 \\
    \addlinespace[3pt]
    \oursblock\oursrow
    BRIDGE (Ours) & ViT-B/16 & BERT-Base &
      \best{\textit{83.04}} & \best{\textit{82.87}} \\
  \end{tabular}
  }%
};

\draw[line width=0.08pt, rounded corners=4pt]
  (tbl.south west) rectangle (tbl.north east);

\draw[line width=0.08pt, rounded corners=4pt, opacity=0.6]
  ([xshift=0.5pt,yshift=0.5pt]tbl.south west)
    rectangle
  ([xshift=0.5pt,yshift=0.5pt]tbl.north east);
\end{tikzpicture}
}

\caption{\textbf{Natural language visual reasoning on NLVR2.}
Accuracy (\%) on the NLVR2 dev and public test set (Test-P) for models with ViT-B/16 and BERT-Base backbones.}
\label{tab:nlvr2}
\end{table}
\section{Ablation Studies}
\label{sec:ablations}

To isolate the effect of the bridge architecture itself, we compare BRIDGE to variants that use the same encoders and training data but differ in how (or whether) cross-modal interaction is performed. As shown in Table~\ref{tab:bridge_arch_ablation}, replacing the bridge with pure late fusion notably degrades performance, while adding even a shallow pooled-only bridge helps. Moving from pooled fusion to hidden-state fusion further improves both TR and IR, and our cross-only interaction design yields the strongest retrieval performance.

\begin{table}[tp]
\centering
\footnotesize
\setlength{\tabcolsep}{4pt}
\renewcommand{\arraystretch}{1.2}

\resizebox{\columnwidth}{!}{%
\begin{tikzpicture}
\node[inner sep=2.5pt] (tbl) {%
  {\tabcomic
  \begin{tabular}{lcccc}
    \multirow{2}{*}{Loss configuration} &
    \multicolumn{2}{c}{MSCOCO (Karpathy 5K)} &
    \multicolumn{2}{c}{VQAv2~\cite{VQA}} \\
    \cmidrule(lr){2-3} \cmidrule(lr){4-5}
    & TR@1 & IR@1 & test-dev & test-std \\
    \midrule
    \oursblock
    \multicolumn{5}{l}{\textbf{Effect of loss components for BRIDGE (ViT-B/16 + BERT-Base)}} \\
    \addlinespace[3pt]
    \oursrow
    InfoNCE only (uni + cross) &
      \textit{79.8} & \textit{65.0} &
      \textit{79.0} & \textit{79.1} \\
    \oursrow
    InfoNCE + ITM (no cycle) &
      \textit{80.8} & \textit{66.4} &
      \textit{79.9} & \textit{80.0} \\
    \oursrow
    InfoNCE + cycle (no ITM) &
      \textit{80.5} & \textit{66.2} &
      \textit{80.1} & \textit{80.2} \\
    \oursrow
    Full model (InfoNCE + ITM + cycle) &
      \best{\textit{81.6}} & \best{\textit{67.5}} &
      \best{\textit{80.6}} & \best{\textit{80.7}} \\
  \end{tabular}
  }%
};

\draw[line width=0.08pt, rounded corners=4pt]
  (tbl.south west) rectangle (tbl.north east);

\draw[line width=0.08pt, rounded corners=4pt, opacity=0.6]
  ([xshift=0.5pt,yshift=0.5pt]tbl.south west)
    rectangle
  ([xshift=0.5pt,yshift=0.5pt]tbl.north east);
\end{tikzpicture}
}

\caption{\textbf{Effect of loss components on MSCOCO retrieval and VQAv2.}
Ablation over the loss functions used to train BRIDGE. All variants use the same ViT-B/16 image encoder,
BERT-Base text encoder, number of interaction layers $Q$, and pretraining data. We report Recall@1 (\%)
on MSCOCO Karpathy 5K and VQAv2 accuracy (\%) on test-dev and test-std.}
\label{tab:bridge_loss_ablation}
\end{table}

\begin{table*}[tp]
\centering
\small
\setlength{\tabcolsep}{5pt}
\renewcommand{\arraystretch}{1.1}

\resizebox{\textwidth}{!}{%
\begin{tikzpicture}
\node[inner sep=2.5pt] (tbl) {%
  {\tabcomic
  \begin{tabular}{lcccccc}
    \multirow{2}{*}{Model} &
    \multicolumn{2}{c}{Cross-Modal Design} &
    \multicolumn{4}{c}{MSCOCO (Karpathy 5K)} \\
    \cmidrule(lr){2-3} \cmidrule(lr){4-7}
    & Interaction Layers & Fusion Type &
    TR@1 & TR@5 & IR@1 & IR@5 \\
    \midrule
    \oursblock
    \multicolumn{7}{l}{\textbf{BRIDGE Architecture Ablations (ViT-B/16 + BERT-Base)}} \\
    \addlinespace[3pt]
    \oursrow
    No-Bridge (late fusion) &
      \textemdash &
      pooled [CLS] only &
      \textit{77.0} & \textit{94.0} & \textit{60.0} & \textit{83.5} \\
    \oursrow
    Pooled-only bridge &
      1 cross block &
      pooled features only &
      \textit{79.5} & \textit{95.5} & \textit{63.0} & \textit{85.0} \\
    \oursrow
    Hidden-state bridge (self+cross) &
      $Q$ layers &
      token-level, self+cross &
      \textit{80.7} & \textit{96.0} & \textit{66.0} & \textit{86.0} \\
    \oursrow
    BRIDGE (cross-only, ours) &
      $Q$ layers &
      token-level, cross-only &
      \best{\textit{81.6}} & \best{\textit{96.6}} &
      \best{\textit{67.5}} & \best{\textit{86.9}} \\
  \end{tabular}
  }%
};

\draw[line width=0.08pt, rounded corners=4pt]
  (tbl.south west) rectangle (tbl.north east);

\draw[line width=0.08pt, rounded corners=4pt, opacity=0.6]
  ([xshift=0.5pt,yshift=0.5pt]tbl.south west)
    rectangle
  ([xshift=0.5pt,yshift=0.5pt]tbl.north east);
\end{tikzpicture}
}

\caption{\textbf{Effect of the bridge architecture on MSCOCO image--text retrieval.}
Ablation over the presence and structure of the BRIDGE interaction layers. All variants use the same ViT-B/16 image encoder, BERT-Base text encoder, and pretraining data; we report Recall@K (\%) on the Karpathy 5K split.}
\label{tab:bridge_arch_ablation}
\end{table*}

\begin{figure*}[tp]
  \centering
  \includegraphics[scale=0.875]{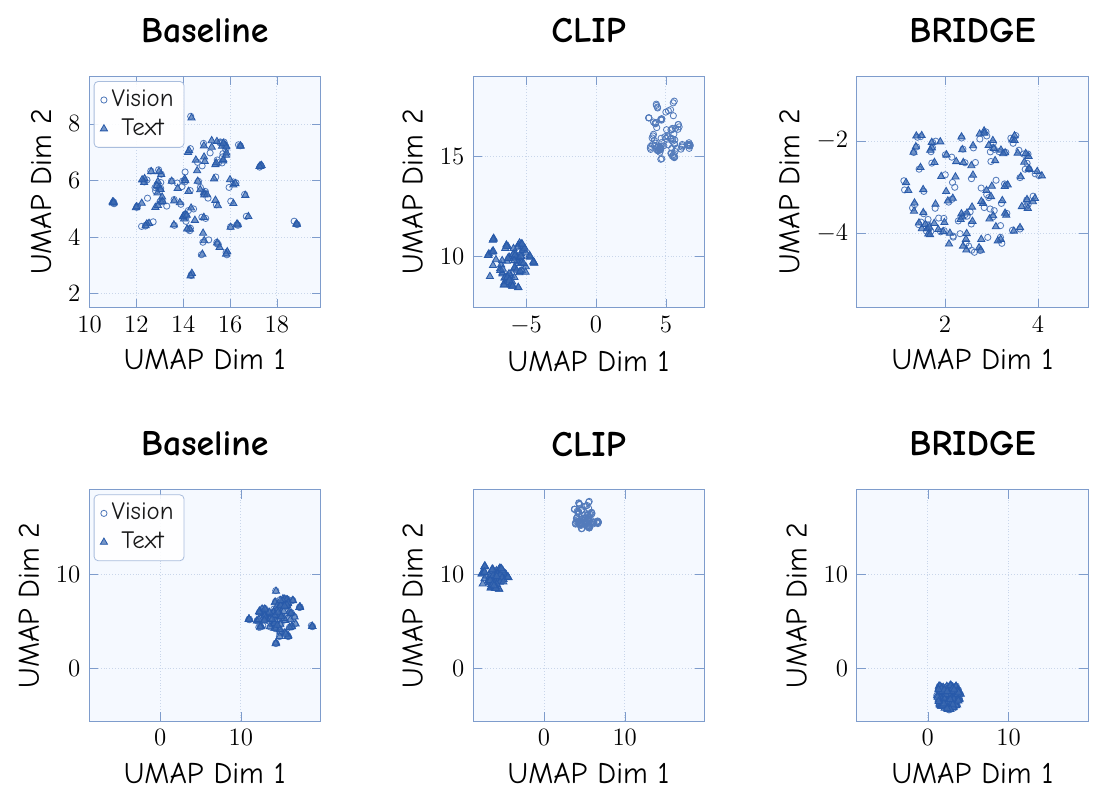}
  \caption{
  UMAP visualization of the joint vision--text embedding space for the baseline encoders, CLIP, and our BRIDGE model. 
  Points denote individual examples, with markers indicating modality (vision vs.\ text). 
  The baseline ecnoders produce diffuse clusters with noticeable separation between modalities, and CLIP, while more compact, still exhibits a clear shift between vision and text regions. 
  BRIDGE instead forms tight, overlapping clusters in which vision and text embeddings for the same examples are nearly co-located, indicating a much smaller modality gap and a more coherent shared representation space. 
  This improved alignment explains the stronger cross-modal retrieval and transfer performance of BRIDGE observed in our experiments. The bottom row contains the same data as the top row, but keeps the axes the same scale.
  }
  \label{fig:umap}
\end{figure*}

To better understand how our model organizes the multimodal representation space, we visualize the learned embeddings using UMAP (Fig.~\ref{fig:umap}). Each panel shows the 2D projection of paired vision and text embeddings for the baseline (standard embeddings of the frozen ViT and BERT encoders), CLIP, and our BRIDGE model. The baseline produces scattered, partially separated clusters, indicating that the two modalities are not well aligned. CLIP tightens the clusters but still exhibits a clear modality gap: vision and text embeddings occupy distinct regions of the space. In contrast, BRIDGE yields compact, highly overlapping clusters where vision and text points for the same examples lie close together. This qualitative behavior matches our quantitative results, suggesting that BRIDGE learns a substantially better aligned and more discriminative joint representation than the alternatives.

We next study where to insert the interaction layers within the encoders. Table~\ref{tab:bridge_placement_ablation} compares using the bridge only in early, middle, or top layers, as well as a staggered configuration. Placing the interaction layers near the top of both encoders consistently performs best, with the staggered variant performing comparably, suggesting that cross-modal fusion is most effective after strong unimodal features have been formed.

\begin{table*}[tp]
\centering
\small
\setlength{\tabcolsep}{5pt}
\renewcommand{\arraystretch}{1.1}

\resizebox{\textwidth}{!}{%
\begin{tikzpicture}
\node[inner sep=2.5pt] (tbl) {%
  {\tabcomic
  \begin{tabular}{lcccccc}
    \multirow{2}{*}{Model} &
    \multicolumn{2}{c}{Interaction Placement} &
    \multicolumn{4}{c}{MSCOCO (Karpathy 5K)} \\
    \cmidrule(lr){2-3} \cmidrule(lr){4-7}
    & Text Encoder & Image Encoder &
    TR@1 & TR@5 & IR@1 & IR@5 \\
    \midrule
    \oursblock
    \multicolumn{7}{l}{\textbf{Effect of Interaction Layer Placement (ViT-B/16 + BERT-Base)}} \\
    \addlinespace[3pt]
    \oursrow
    Early-only &
      lowest $Q$ layers &
      lowest $Q$ layers &
      \textit{80.2} & \textit{96.0} & \textit{66.2} & \textit{86.0} \\
    \oursrow
    Middle-only &
      middle $Q$ layers &
      middle $Q$ layers &
      \textit{81.0} & \textit{96.3} & \textit{66.9} & \textit{86.3} \\
    \oursrow
    Late-only (top) &
      highest $Q$ layers &
      highest $Q$ layers &
      \best{\textit{81.6}} & \best{\textit{96.6}} &
      \best{\textit{67.5}} & \best{\textit{86.9}} \\
    \oursrow
    Staggered (uniform) &
      uniformly spaced $Q$ layers &
      uniformly spaced $Q$ layers &
      \textit{81.4} & \textit{96.5} & \textit{67.3} & \textit{86.7} \\
  \end{tabular}
  }%
};

\draw[line width=0.08pt, rounded corners=4pt]
  (tbl.south west) rectangle (tbl.north east);

\draw[line width=0.08pt, rounded corners=4pt, opacity=0.6]
  ([xshift=0.5pt,yshift=0.5pt]tbl.south west)
    rectangle
  ([xshift=0.5pt,yshift=0.5pt]tbl.north east);
\end{tikzpicture}
}

\caption{\textbf{Effect of interaction-layer placement on MSCOCO image--text retrieval.}
Ablation over which encoder layers are selected as cross-modal interaction layers. All variants use the same ViT-B/16 image encoder, BERT-Base text encoder, number of interaction layers $Q$, and pretraining data; we report Recall@K (\%) on the Karpathy 5K split.}
\label{tab:bridge_placement_ablation}
\end{table*}

Finally, we ablate the training objectives used for BRIDGE. Table~\ref{tab:bridge_loss_ablation} reports MSCOCO retrieval and VQAv2 performance when using only unimodal and cross-modal InfoNCE, adding ITM, adding cycle-consistency, or combining all three. While InfoNCE alone already gives strong results, both ITM and the cycle-consistent cross-attention loss contribute additional gains, and the full loss cocktail yields the best retrieval and VQA accuracy.

\section{Limitations}
\label{sec:limitations}
While BRIDGE improves cross-modal alignment and achieves SoTA performance on our experiments, our study has several limitations. First, our backbone encoder models are relatively small, and the number of interaction layers is kept small due to computational budget, so our results may understate the potential gains at larger scales. Second, we evaluate on a focused set of benchmarks; instruction-tuned MLLMs such as LLaVA and related systems, as well as large contrastive or captioning models like CLIP and BLIP, may still outperform BRIDGE on open-ended dialogue or heavily instruction-following settings where they are specifically optimized. We view BRIDGE as complementary to these approaches, providing a modular fusion mechanism that  preserves efficient unimodal inference and can be paired with existing decoders.

\newpage \
\section{Conclusion}
\label{sec:conclusion}
We introduced BRIDGE, a lightweight fusion module that directly aligns the hidden states of vision and text encoders using a small number of cross-only, bidirectional attention layers. The design of BRIDGE preserves bi-encoder efficiency for retrieval, while also enabling token-level interaction that improves multimodal representation. BRIDGE allows for a simple staged training approach that can be adapted for other VLMs and MLLMs. Across MSCOCO and Flickr30K retrieval, VQAv2, and NLVR2, BRIDGE delivers consistent gains over comparable backbones. Ablations show that hidden-state fusion (rather than pooled-only fusion), late placement of interaction layers, and the full combination of losses are key contributors to performance.

\FloatBarrier
{
    \small
    \bibliographystyle{ieeenat_fullname}
    \bibliography{main}
}


\end{document}